\documentclass[aps,superscriptaddress,twocolumn,longbibliography]{revtex4-1}
\pdfoutput=1
\usepackage{amsmath}
\usepackage{amssymb}
\usepackage{graphicx}
\usepackage{xcolor}
\usepackage{braket}
\usepackage{verbatim}
\usepackage{marginnote}
\usepackage{mathtools}
\usepackage{hyperref}

\newcommand{\beq}{\begin{equation}}
\newcommand{\eeq}{\end{equation}}
\newcommand{\beqa}{\begin{eqnarray}}
\newcommand{\eeqa}{\end{eqnarray}}
\newcommand{\ben}{\begin{enumerate}}
\newcommand{\een}{\end{enumerate}}

\begin{document}

\title{Supervised Learning With Quantum-Inspired Tensor Networks}

\author{E.\ Miles Stoudenmire}
\affiliation{Perimeter Institute for Theoretical Physics, Waterloo, Ontario, N2L 2Y5, Canada}
\affiliation{Department of Physics and Astronomy, University of California, Irvine, CA 92697-4575 USA}

\author{David J.\ Schwab}
\affiliation{Dept.\ of Physics, Northwestern University, Evanston, IL}

\date{\today}

\begin{abstract}
Tensor networks are efficient representations of high-dimensional
tensors which have been very successful for physics and mathematics applications.
We demonstrate how algorithms for optimizing such networks 
can be adapted to supervised learning tasks by using matrix product states 
(tensor trains) to parameterize models for classifying images. For the
MNIST data set we obtain less than $1\%$ test set classification error. We 
discuss how the tensor network form imparts additional structure to the 
learned model and suggest a possible generative interpretation.
\end{abstract}

%
%
%
%
%
%

\maketitle

\section{Introduction}


\begin{figure}[b]
\includegraphics[width=\columnwidth]{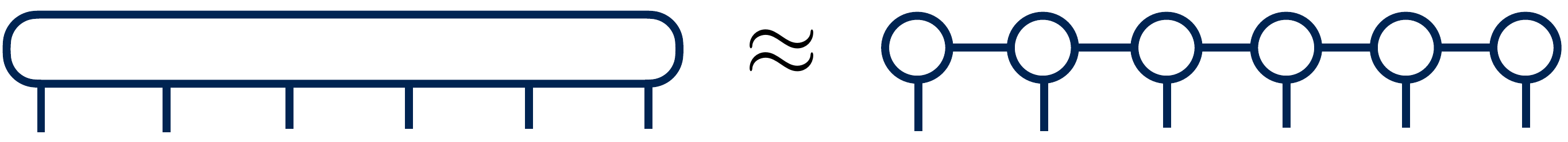}
\caption{The matrix product state (MPS) decomposition, also known as a tensor train. Lines
represent tensor indices and connecting two lines implies summation. For an introduction to this graphical 
tensor notation see Appendix~\ref{appendix:graphical}.}
\label{fig:mps}
\end{figure}

The connection between machine learning and statistical physics has 
long been appreciated \cite{Hopfield:1982,Amit:1985,Derrida:1987,Amit:1987,Seung:1992,Engel:2001,Malzahn:2005,Hinton:2006,Mezard:2009}, but deeper relationships continue to be uncovered.
For example, techniques used to pre-train
neural networks \cite{Hinton:2006} have more recently been interpreted 
in terms of the renormalization group \cite{Mehta:2014}.
In the other direction there has been a sharp increase in applications of machine 
learning to chemistry, material science, and condensed matter physics 
\cite{Fischer:2006,Hautier:2010,Rupp:2012,Saad:2012,Snyder:2012,Pilania:2013,Arsenault:2014,Amin:2016aa,Carrasquilla:2017},
which are sources of highly-structured data and could be a good testing ground for machine learning techniques.

A recent trend in both physics and machine learning is an appreciation for 
the power of tensor methods. In machine learning, tensor decompositions can be used
to solve non-convex optimization tasks \cite{Anandkumar:2014,Anandkumar:2014a} and 
make progress on many 
other important problems \cite{Phan:2010,Bengua:2015a,Novikov:2015}, while in physics, 
great strides have been made in manipulating large vectors arising in quantum mechanics
by decomposing them as \emph{tensor networks} \cite{Bridgeman:2016,Schollwoeck:2011,Evenbly:2011g}. 
The most successful types of 
tensor networks avoid the curse of dimensionality by incorporating only low-order tensors, 
yet accurately reproduce very high-order tensors through a particular geometry of 
tensor contractions \cite{Evenbly:2011g}.

Another context where very large vectors arise is in non-linear kernel learning,
where  input vectors $\mathbf{x}$ are mapped
into a higher dimensional space via a feature map $\Phi(\mathbf{x})$ before
being classified by a decision function 
\begin{equation}
f(\mathbf{x}) = W \cdot \Phi(\mathbf{x}) \ . \label{eqn:1classmodel}
\end{equation}
The feature vector $\Phi(\mathbf{x})$ and weight vector
$W$ can be exponentially large or even infinite. One approach to
deal with such large vectors is the well-known
kernel trick, which only requires working with scalar products of 
feature vectors, allowing these vectors to be defined only implicitly \cite{Muller:2001}.

In what follows we propose a rather different approach. For certain
learning tasks and a specific class of feature map $\Phi$, we find 
the optimal weight vector $W$ can be approximated 
as a tensor network, that is, as a contracted sequence of low-order
tensors.
Representing W as a tensor network and optimizing it directly (without passing to the dual representation) has many interesting 
consequences. Training the model scales linearly in the training set size; the cost for evaluating an input is independent 
of training set size. Tensor networks are also adaptive: dimensions of tensor indices internal to the network grow and shrink 
during training to concentrate resources on the particular correlations within the data most useful for learning. 
The tensor network form of W presents opportunities to extract information hidden within the trained model and to accelerate 
training by using techniques such as optimizing different internal tensors in parallel \cite{Stoudenmire:2013p}. 
Finally, the tensor network form is an additional type of regularization beyond the choice of feature map, 
and could have interesting consequences for generalization.

One of the best understood types of tensor networks is 
the matrix product state~\cite{Ostlund:1995,Schollwoeck:2011}, also known as 
the tensor train decomposition~\cite{Oseledets:2011}. 
Matrix product states (MPS) have been very useful for studying quantum systems, and 
have recently been proposed for machine learning applications
such as learning features of images~\cite{Bengua:2015a} and compressing the weight layers
of neural networks \cite{Novikov:2015}.
Though MPS are best suited for describing one-dimensional
systems, they are powerful enough to be applied to higher-dimensional
systems as well. 

There has been intense research into generalizations of MPS better suited for higher
dimensions and critical systems \cite{Verstraete:2004p,Vidal:2007,Evenbly:2009}.
Though our proposed approach could generalize to these other types of tensor networks,
as a proof of principle
we will only consider the MPS decomposition in what follows.
The MPS decomposition approximates an order-N tensor by a contracted chain
of N lower-order tensors shown in Fig.~\ref{fig:mps}. (Throughout we will use
tensor diagram notation; for a brief review see Appendix~\ref{appendix:graphical}.)

Representing the weights $W$ of Eq.~(\ref{eqn:1classmodel})
as an MPS allows us to efficiently optimize these weights and adaptively change their number 
by varying $W$ locally a few tensors at a time, in close analogy to the density matrix 
renormalization group algorithm used in physics \cite{White:1992,Schollwoeck:2011}. 
Similar alternating least squares methods for tensor trains have also been explored in 
applied mathematics  \cite{Holtz:2012}.

This paper is organized as follows: we propose our general approach 
then describe an algorithm for optimizing the weight vector $W$ in MPS form.
We test our approach, both on the MNIST handwritten digit set and on two-dimensional
toy data to better understand the role of the local feature-space dimension $d$.
Finally, we discuss the class of functions realized by our proposed models as well 
as a possible generative interpretation.

Those wishing to reproduce our results can find sample codes based on the ITensor library \cite{ITensor}
at:
\href{https://github.com/emstoudenmire/TNML}{https://github.com/emstoudenmire/TNML}

\section{Encoding Input Data \label{sec:encoding}}

The most successful use of tensor networks in physics so far has been
in quantum mechanics, where 
combining $N$ independent systems corresponds to taking the tensor product
of their individual state vectors. With the goal of applying similar tensor networks
to machine learning,  
we choose a feature map of the form
\begin{equation}
\Phi^{s_1 s_2 \cdots s_N}(\mathbf{x}) = \phi^{s_1}(x_1) \otimes \phi^{s_2}(x_2) \otimes \cdots \phi^{s_N}(x_N) \ .
\label{eqn:tensor_prod}
\end{equation}
The tensor $\Phi^{s_1 s_2 \cdots s_N}$ is the tensor product of the same local feature map $\phi^{s_j}(x_j)$ applied to 
each input $x_j$, where the indices $s_j$ run from $1$ to $d$; the value $d$ is known as the local 
dimension. Thus each $x_j$ is mapped to a $d$-dimensional vector, which we require to have unit norm;
this implies each $\Phi(\mathbf{x})$ also has unit norm.

\begin{figure}[t]
\includegraphics[width=0.65\columnwidth]{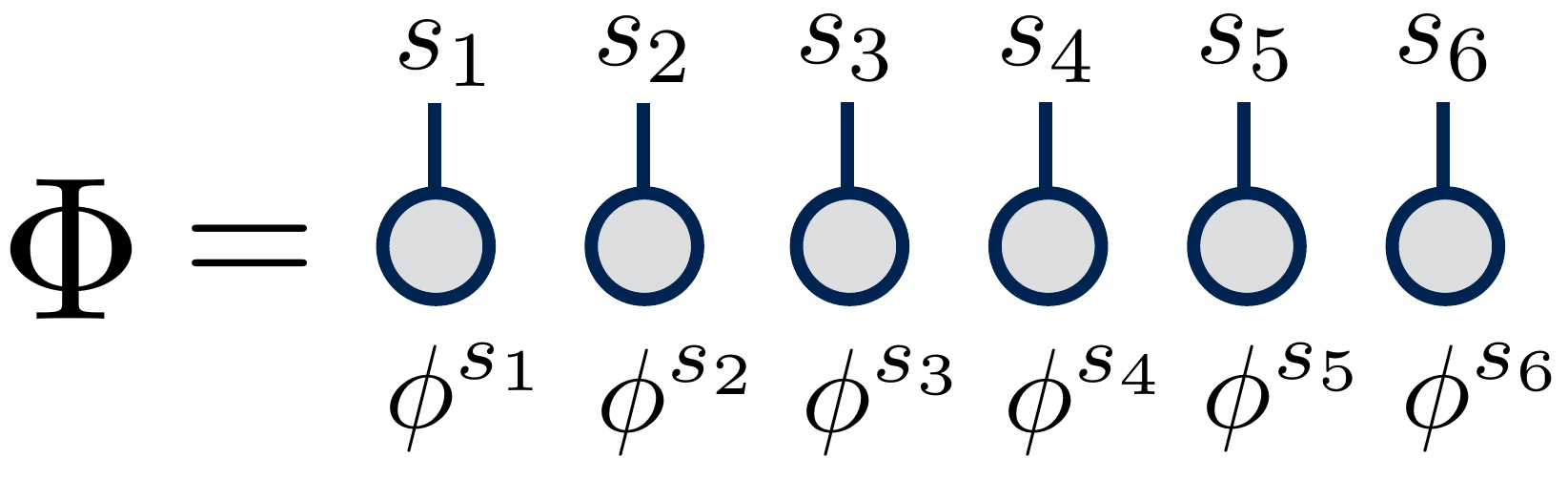}
\caption{Input data is mapped to a normalized order N tensor with a trivial (rank 1) product structure.}
\label{fig:input}
\end{figure}

The full feature map $\Phi(\mathbf{x})$ can be viewed
as a vector in a $d^N$-dimensional space or as an order-$N$ tensor. 
The tensor diagram for $\Phi(\mathbf{x})$ is shown in Fig.~\ref{fig:input}. This type of tensor is said be 
rank-1 since it is manifestly the product of $N$ order-1 tensors. 
In physics terms, $\Phi(\mathbf{x})$ has the same structure as a product state or unentangled wavefunction.

\begin{figure}[t]
\includegraphics[width=0.65\columnwidth]{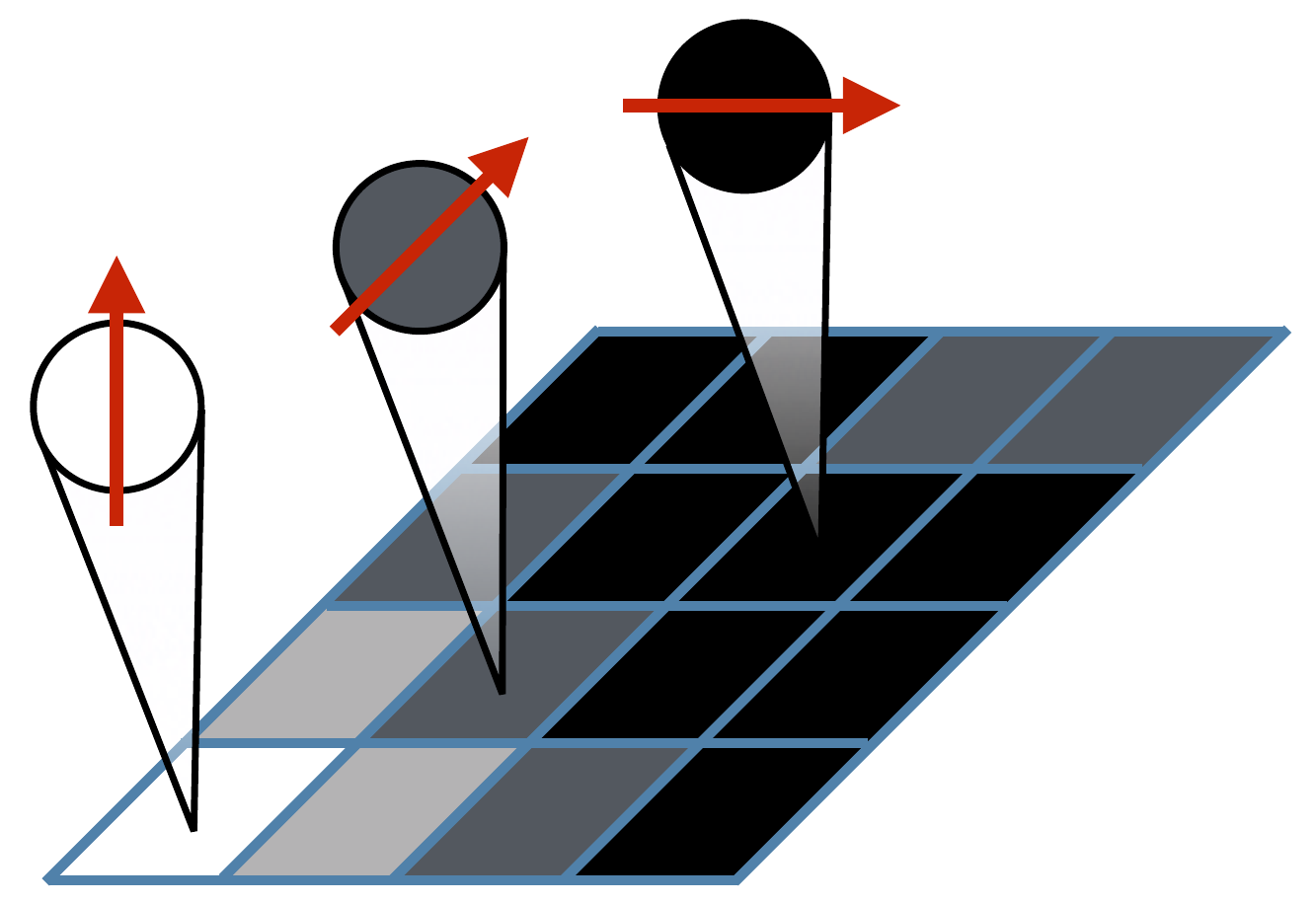}
\caption{For the case of a grayscale image and $d=2$, each pixel value is mapped to a normalized 
two-component vector. The full image is mapped to the tensor product of all the local pixel vectors as 
shown in Fig.~\ref{fig:input}.}
\label{fig:encoding}
\end{figure}

For a concrete example of this type of feature map, consider  inputs which are grayscale images
with $N$ pixels, where each pixel value ranges from 0.0 for white to 1.0 for black. 
If the grayscale pixel value of the $j^\text{th}$ pixel is \mbox{$x_j \in [0,1]$},
a simple choice for the local feature map $\phi^{s_j}(x_j)$ is
\begin{equation}
\phi^{s_j}(x_j) = \left[ \cos{\left(\frac{\pi}{2} x_j\right)},\ \sin{\left(\frac{\pi}{2} x_j\right)} \right]
\label{eqn:gray_mapping}
\end{equation}
and is illustrated in Fig.~\ref{fig:encoding}.
The full image is represented as a tensor product of these local vectors.
From a physics perspective, $\phi^{s_j}$ is the normalized wavefunction of a single qubit where the 
``up'' state corresponds to a white pixel, the ``down'' state to a black pixel, and a superposition
corresponds to a gray pixel.


While our choice of feature map $\Phi(\mathbf{x})$ was originally motivated from a physics perspective,
in machine learning terms, the feature map Eq.~(\ref{eqn:tensor_prod}) defines
a kernel which is the product of $N$ local kernels, one for each component $x_j$ of the input data. 
Kernels of this type have been discussed previously \cite[p. 193]{Vapnik:2000} and have been
argued to be useful for data where no relationship is assumed between different components of the input
vector prior to learning \cite{Waegeman:2012}.

Though we will use only the local feature map Eq.~(\ref{eqn:gray_mapping}) in our MNIST experiment below,
it would be interesting to try other local maps and to understand better the role they play in the performance
of the model and the cost of optimizing the model.


\section{Multiple Label Classification \label{sec:tensors}}

In what follows we are interested in multi-class learning, for which we choose a 
``one-versus-all'' strategy, which we take to mean generalizing the decision function 
Eq.~(\ref{eqn:model}) to a set of functions indexed by a label $\ell$
\begin{equation}
f^\ell(\mathbf{x}) = W^\ell \cdot \Phi(\mathbf{x}) \label{eqn:model}
\end{equation}
and classifying an input $\mathbf{x}$ by choosing the label $\ell$ for which
$|f^\ell(\mathbf{x})|$ is largest.

\begin{figure}[t]
\includegraphics[width=0.85\columnwidth]{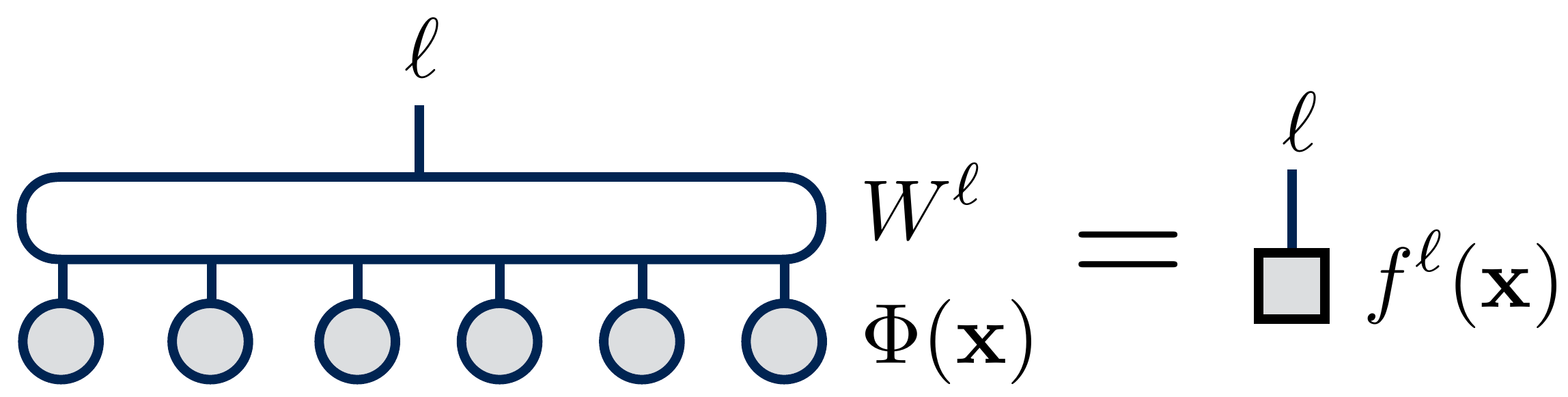}
\caption{The overlap of the weight tensor $W^\ell$ with a specific input vector $\Phi(\mathbf{x})$
defines the decision function $f^\ell(\mathbf{x})$. 
The label $\ell$ for which $f^\ell(\mathbf{x})$ has maximum magnitude is the predicted label for $\mathbf{x}$.}
\label{fig:overlap}
\end{figure}

Since we apply the same feature map $\Phi$ to all input data, the only quantity that
depends on the label $\ell$ is the weight vector $W^\ell$. 
Though one can view $W^\ell$ as a collection of vectors labeled by $\ell$,
we will prefer to view $W^\ell$ as an order $N+1$ tensor where $\ell$ is 
a tensor index and $f^\ell(\mathbf{x})$ is a function mapping inputs to the space of
labels. The tensor diagram for evaluating $f^\ell(\mathbf{x})$ for a particular
input is depicted in Fig.~\ref{fig:overlap}.



\section{MPS Approximation \label{sec:mps}}

Because the weight tensor $W^\ell_{s_1 s_2 \cdots s_N}$ has $N_L \cdot d^N$
components, where $N_L$ is the number of labels, we need a way to regularize and optimize this
tensor efficiently.
The strategy we will use is to represent this high-order tensor as a \emph{tensor network},
that is, as the contracted product of lower-order tensors. 

A tensor network approximates the exponentially large
set of components of a high-order tensor in terms of a much smaller
set of parameters whose number grows only polynomially in the size of the input space.
Various tensor network approximations impose different assumptions, or implicit priors, about the pattern
of correlation of the local indices when viewing the original tensor as a distribution.
For example, a MERA network can explicitly model power-law decaying correlations while a 
matrix product state (MPS) has exponentially decaying correlations \cite{Evenbly:2011g}. Yet an MPS
can still approximate power-law decays over quite long distances.

For the rest of this work, we will approximate $W^\ell$ as an MPS,
which have the key advantage that methods for manipulating
and optimizing them are well understood and highly efficient.
Although MPS are best suited for approximating tensors with a one-dimensional pattern of 
correlations, they can also be a powerful approach for decomposing tensors 
with two-dimensional correlations as well \cite{Stoudenmire:2012a}.

\begin{figure}[t]
\includegraphics[width=\columnwidth]{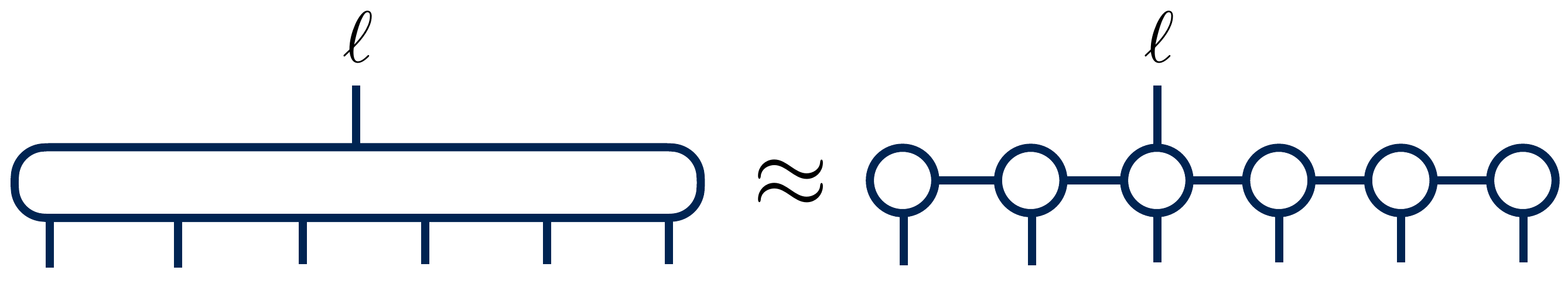}
\caption{Approximation of the weight tensor $W^\ell$ by a matrix product state. The label
index $\ell$ is placed arbitrarily on one of the MPS tensors but can be moved to other locations.}
\label{fig:classifier_mps}
\end{figure}

An MPS decomposition of the weight tensor $W^\ell$ has the form
\begin{equation}
W^\ell_{s_1 s_2 \cdots s_N} = \sum_{\{\alpha\}} A^{\alpha_1}_{s_1} A^{\alpha_1 \alpha_2}_{s_2} \cdots A^{\ell; \alpha_j \alpha_{j+1}}_{s_j} \cdots A^{\alpha_{N-1}}_{s_N} \label{eqn:mps}
\end{equation}
and is illustrated in Fig.~\ref{fig:classifier_mps}.
Each ``virtual'' index $\alpha_j$ has a dimension $m$ which
is known as the \emph{bond dimension} and is the (hyper) parameter controlling the MPS approximation.
For sufficiently large $m$ an MPS can represent any tensor \cite{Verstraete:2004}.
The name matrix product state refers to the fact that any specific 
component of the full tensor $W^\ell_{s_1 s_2 \cdots s_N}$ can be recovered efficiently
by summing over the $\{\alpha_j\}$ indices from left to right via a sequence of matrix products.

In the above decomposition, the label index $\ell$ was arbitrarily placed on the
$j^\text{th}$ tensor, but this index can be moved to any other tensor of the MPS without changing the overall
$W^\ell$ tensor it represents. To do this, one contracts the $j^\text{th}$ tensor with one of its neighbors, then decomposes 
this larger tensor using a singular value decomposition such that $\ell$ now belongs to the neighboring
tensor---see Fig.~\ref{fig:sweeping2}(b).

In typical physics applications the MPS bond dimension 
$m$ can range from 10  to 10,000 or even more;
for the most challenging physics systems one wants to allow as large 
a bond dimension as possible since a larger dimension means more accuracy.
However, when using MPS in a machine learning setting, the bond dimension
controls the number of parameters of the model. So in contrast
to physics, taking too large a bond dimension might not be desirable
as it could lead to overfitting.


\section{Sweeping Algorithm for Optimizing Weights \label{sec:algorithm}}

Inspired by the very successful density matrix renormalization group (DMRG) algorithm
developed in physics, here we propose
a similar algorithm which ``sweeps'' back and forth along an MPS, iteratively minimizing
the cost function defining the classification task.

For concreteness, let us choose to optimize the quadratic cost
\begin{align}
C = \frac{1}{2} \sum_{n=1}^{N_T} \sum_{\ell} (f^\ell(\mathbf{x}_n) - \delta^\ell_{L_n})^2
\end{align}
where $n$ runs over the $N_T$ training inputs and $L_n$ is the known correct label for training input $n$.
(The symbol $\delta^{\ell}_{L_n} = 1$ if $\ell$ equals $L_n$ and zero otherwise, and represents the ideal output
of the function $f^\ell$ for the input $\mathbf{x}_n$.)

Our strategy for reducing this cost function will be to vary only two neighboring MPS tensors
at a time within the approximation Eq.~(\ref{eqn:mps}). We could conceivably just vary one at a 
time but as will become clear, varying two tensors leads to a straightforward method for adaptively
changing the MPS bond dimension.

\begin{figure}[t]
\includegraphics[width=\columnwidth]{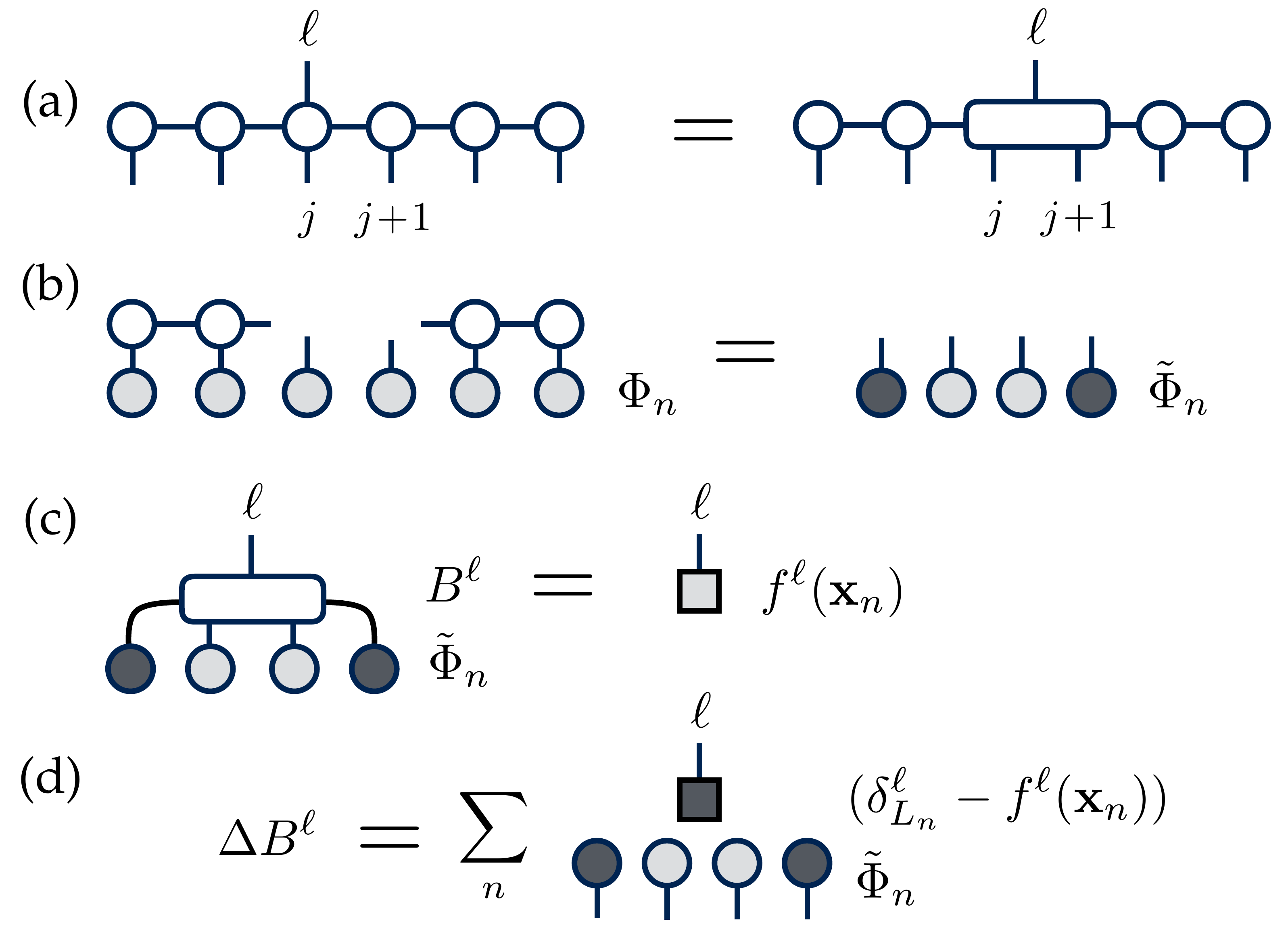}
\caption{Steps leading to computing the gradient of the bond tensor $B^\ell$ at bond $j$: (a) forming the 
bond tensor; (b) projecting a training input into the ``MPS basis'' at bond $j$; (c)~computing the decision
function in terms of a projected input; (d) the gradient correction to $B^\ell$. The dark shaded circular tensors
 in step (b) are ``effective features'' formed from $m$ different linear combinations of many original
features.}
\label{fig:sweeping1}
\end{figure}

Say we want to improve the tensors at sites $j$ and $j+1$ which share the $j^\text{th}$ bond. 
Assume we have moved the label index $\ell$ to the $j^\text{th}$ MPS tensor. First
we combine the MPS tensors $A^\ell_{s_j}$ and $A_{s_{j+1}}$ into a single ``bond tensor''
$B^{\alpha_{j-1} \ell \alpha_{j+1}}_{s_j s_{j+1}}$ by contracting over the index $\alpha_j$ as shown in Fig.~\ref{fig:sweeping1}(a).

Next we compute the derivative of the cost function $C$ with respect to the bond tensor $B^\ell$ 
in order to update it using a gradient descent step.
Because the rest of the MPS tensors are kept fixed, let us show that to compute the
gradient it suffices to feed, or project, each input $\mathbf{x}_n$ through the fixed ``wings'' of 
the MPS as shown on the left-hand side of Fig.~\ref{fig:sweeping1}(b). Doing so produces the projected, 
four-index version of the input $\tilde{\Phi}_n$ shown on the right-hand of Fig.~\ref{fig:sweeping1}(b). 
The current decision function can be efficiently computed from this projected input $\tilde{\Phi}_n$ and 
the current bond tensor $B^\ell$ as 
\begin{align}
f^\ell(\mathbf{x}_n) = \sum_{\alpha_{j-1} \alpha_{j+1}}\sum_{s_j s_{j+1}} B^{\alpha_{j-1} \ell \alpha_{j+1}}_{s_j s_{j+1}} (\tilde{\Phi}_n)_{\alpha_{j-1} \ell \alpha_{j+1}}^{s_j s_{j+1}}
\end{align}
or as illustrated in Fig.~\ref{fig:sweeping1}(c). 
Thus the leading-order update to the tensor $B^\ell$ can be computed as
\begin{align}
\Delta B^\ell & \stackrel{\text{def}}{=} - \frac{\partial C}{\partial B^\ell}  \\
& = \sum_{n=1}^{N_T} \sum_{\ell'} (\delta^{\ell'}_{L_n}-f^{\ell'}(\mathbf{x}_n)) \frac{\partial f^{\ell'}(\mathbf{x}_n)}{\partial B^\ell} \\
& = \sum_{n=1}^{N_T} (\delta^\ell_{L_n}-f^\ell(\mathbf{x}_n)) \tilde{\Phi}_n \ .
\end{align}
Note that last expression above is a tensor with the same index structure 
as $B^\ell$ as shown in Fig.~\ref{fig:sweeping1}(d).

\begin{figure}[t]
\includegraphics[width=\columnwidth]{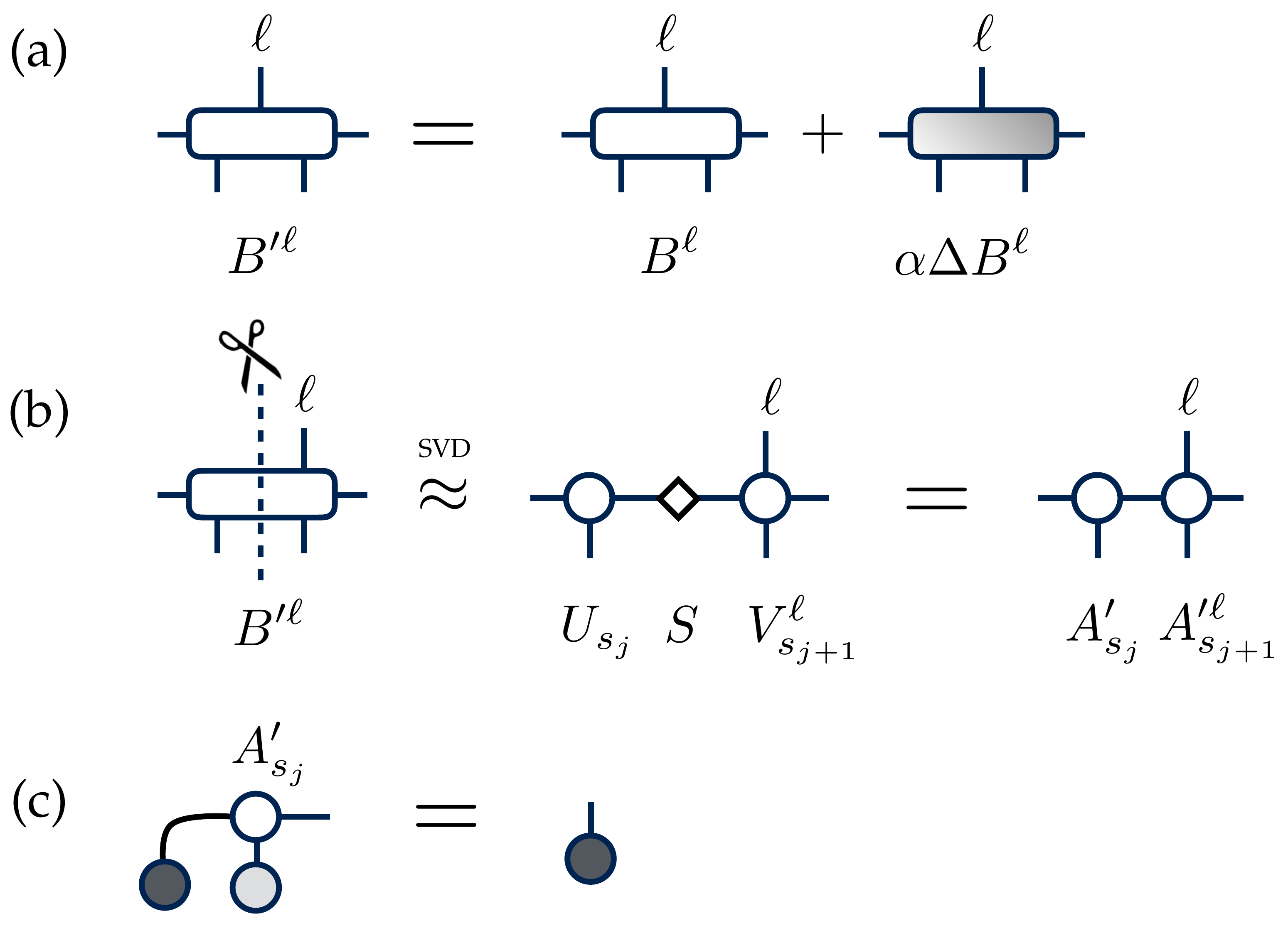}
\caption{Update (a) of bond tensor $B^\ell$, (b) restoration of MPS form, and (c) advancing
a projected training input before optimizing the tensors at the next bond.}
\label{fig:sweeping2}
\end{figure}

Assuming we have computed the gradient, we use it to compute a small update to
$B^\ell$, replacing it with $B^\ell + \alpha \Delta B^\ell$ as shown in Fig.~\ref{fig:sweeping2}(a), where
$\alpha$ is a small empirical paramater used to control convergence. (Note that for this step one could
also use the conjugate gradient method to improve the performance.)
Having obtained our improved $B^\ell$, we must decompose it back into separate
MPS tensors to maintain efficiency and apply our algorithm to the next bond. Assume the next bond we
want to optimize is the one to the right (bond $j+1$). Then we can compute a singular value 
decomposition (SVD) of $B^\ell$, treating it as a matrix with a collective row index $(\alpha_{j-1}, s_j)$ 
and collective column index $(\ell,\alpha_{j+1},s_{j+1})$ as shown in Fig.~\ref{fig:sweeping2}(b).
Computing the SVD this way restores the MPS form, but with the $\ell$ index moved to the 
tensor on site $j+1$. If the SVD of $B^\ell$ is given by
\begin{equation}
B^{\alpha_{j-1} \ell \alpha_{j+1}}_{s_j s_{j+1}} = \sum_{\alpha^\prime_j \alpha_j} U^{\alpha_{j-1}}_{s_j \alpha^\prime_j} S^{\alpha^\prime_j}\,_{\alpha_j} 
V^{\alpha_j \ell \alpha_{j+1}}_{s_{j+1}} \ ,
\end{equation}
then to proceed to the next step we define the new MPS tensor at site $j$ to be 
$A^\prime_{s_j} = U_{s_j}$ and the new tensor at site $j+1$ to be $A^{\prime \ell}_{s_{j+1}} = S V^{\ell}_{s_{j+1}}$ 
where a matrix multiplication over the suppressed $\alpha$ indices is implied. 
Crucially at this point, only the $m$ largest singular values in $S$ are kept and the rest are truncated (along
with the corresponding columns of $U$ and $V^\dagger$) in order
to control the computational cost of the algorithm.
Such a truncation is guaranteed to produce an optimal approximation of the tensor $B^\ell$; furthermore if all of the MPS tensors to the left and right of $B^\ell$ are formed 
from (possibly truncated) unitary matrices similar to the definition of $A^\prime_{s_j}$ above, then 
the optimality of the truncation of $B^\ell$ applies globally to the entire MPS as well. For further
background reading on these technical aspects of MPS, see Refs.~\onlinecite{Schollwoeck:2011}~and~\onlinecite{Cichocki:2014b}.

Finally, when proceeding to the next bond, it would be inefficient to fully project each training input
over again into the configuration in Fig.~\ref{fig:sweeping1}(b). Instead it is only necessary to advance the projection
by one site using the MPS tensor set from a unitary matrix after the SVD as shown in Fig.~\ref{fig:sweeping2}(c). 
This allows the cost of each local step of the algorithm to remain independent of the size of the input space,
making the total algorithm scale only linearly with input space size.

The above algorithm highlights a key advantage of MPS and tensor networks 
relevant to machine learning applications. Following the SVD of the improved bond tensor $B^{\prime \ell}$,
the dimension of the new MPS bond can be chosen \emph{adaptively} 
based on number of large singular values (defined by a threshold chosen in advance).
Thus the MPS form of $W^\ell$ can be compressed as much as possible, and by different amounts on each bond,
while still ensuring an optimal decision function.

The scaling of the above algorithm is $d^3\, m^3 \, N\,N_L\, N_T$, where recall $m$ is the MPS bond dimension;
$N$ the number of input components; $N_L$ the number of labels; and $N_T$ the number of training inputs. 
In practice, the cost is dominated by the large number of training inputs $N_T$, so it would be very
desirable to reduce this cost. One solution could be to use stochastic gradient descent, but
while our experiments at blending this approach with the MPS sweeping algorithm often reached single-digit
classification errors, we could not match the accuracy of the full gradient.
Mixing stochastic gradient with MPS sweeping thus appears to be non-trivial but we believe
it is a promising direction for further research.

Finally, we note that a related optimization algorithm was proposed for hidden Markov models in Ref.~\onlinecite{Rolfe:2010}.
However, in place of our SVD above, Ref.~\onlinecite{Rolfe:2010} uses a non-negative
matrix factorization. In a setting where negative weights are allowed, the SVD is the optimal choice because 
it minimizes the distance between the original tensor and product of factorized tensors. Furthermore, our framework
for sharing weights across multiple labels and our use of local feature maps has interesting implications for training performance
and for generalization.

\section{MNIST Handwritten Digit Test \label{sec:mnist}}

To test the tensor network approach on a realistic task, we used
the MNIST data set, which consists of grayscale images of the digits 
zero through nine \cite{MNIST}. The calculations were implemented using the ITensor library \cite{ITensor}.
Each image was originally $28\times28$ pixels, which we
scaled down to $14\times14$ by averaging clusters of four pixels; otherwise
we performed no further modifications to the training or test sets.
Working with smaller images reduced the time needed for training,
with the tradeoff being that less information was available for learning.

\begin{figure}[t]
\includegraphics[width=0.8\columnwidth]{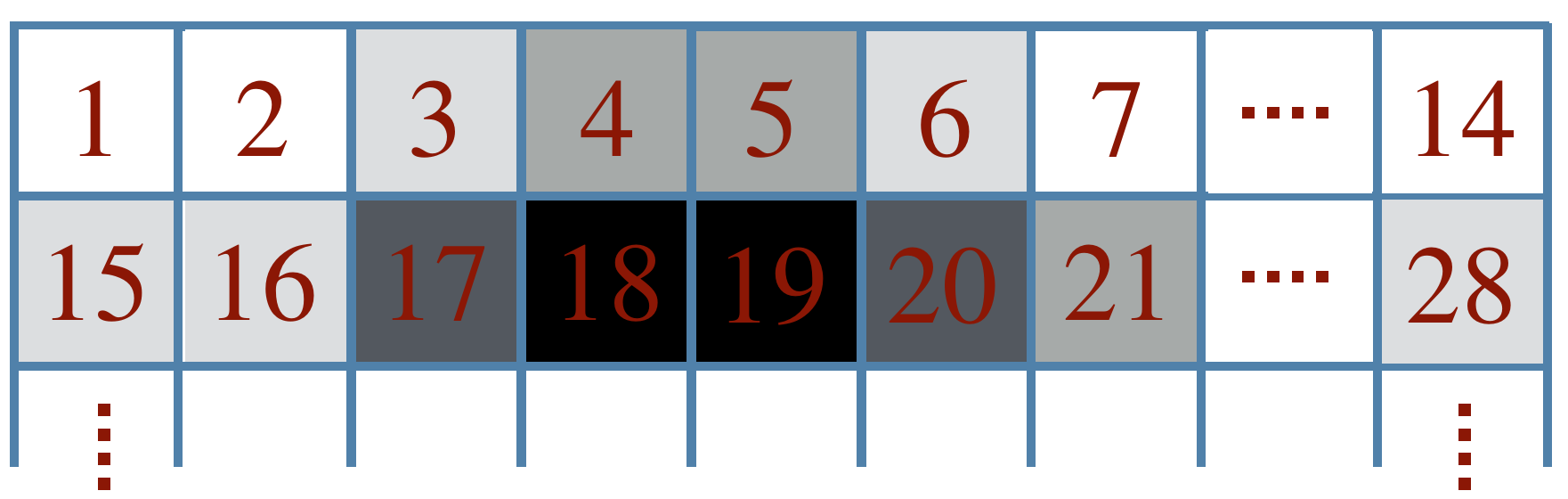}
\caption{One-dimensional ordering of pixels used to train MPS classifiers for the MNIST data set 
(after shrinking images to $14\times14$ pixels).}
\label{fig:ordering}
\end{figure}

To approximate the classifier tensors as MPS, one must choose a one-dimensional
ordering of the local indices $s_1, s_2, \ldots, s_N$. 
We chose a ``zig-zag'' ordering shown in Fig.~\ref{fig:ordering}, which on average keeps
spatially neighboring pixels as close to each other as possible along the
one-dimensional MPS path. We then mapped each grayscale image $\mathbf{x}$ to a tensor 
$\Phi(\mathbf{x})$ using the local map Eq.~(\ref{eqn:gray_mapping}).


Using the sweeping algorithm in Section~\ref{sec:algorithm} to train the weights, 
we found the algorithm quickly converged in the number of passes, or sweeps
over the MPS. Typically only two or three sweeps were needed to see good convergence,
with test error rates changing only hundreths of a percent thereafter.

Test error rates also decreased rapidly with the maximum MPS bond dimension $m$. For
$m=10$ we found both a training and test error of about 5\%; 
for $m=20$ the error dropped to only 2\%.
The largest bond dimension we tried was $m=120$, where
after three sweeps we obtained a test error of 0.97\% (97 misclassified
images out of the test set of 10,000 images); the
training set error was 0.05\% or 32 misclassified images.


\section{Two-Dimensional Toy Model \label{sec:toy}}

To better understand the modeling power and regularization properties of the class of models 
presented in Sections~\ref{sec:encoding}~and~\ref{sec:tensors},
consider a family of toy models where the input space is two-dimensional ($N=2$).
The hidden distribution we want to learn consists of two distributions,
$P_A(x_1,x_2)$ and $P_B(x_1,x_2)$, from which we generate training data points labeled
$A$ or $B$ respectively. For simplicity we only consider the square region
$x_1 \in [0,1]$ and $x_2 \in [0,1]$.

To train the model, each training point $(x_1,x_2)$ is mapped to a tensor
\begin{equation}
\Phi(x_1,x_2) = \phi^{s_1}(x_1) \otimes \phi^{s_2}(x_2)
\end{equation}
and the full weight tensors $W^\ell_{s_1 s_2}$ for $\ell \in \{A,B\}$ are optimized
directly using gradient descent.

When selecting a model, our main control parameter is the dimension $d$ 
of the local indices $s_1$ and $s_2$.
For the case $d=2$, the local feature map is chosen as in Eq.~\ref{eqn:gray_mapping}.
For $d>2$ we generalize $\phi^{s_j}(x_j)$ to be a normalized $d$-component vector
as described in Appendix~\ref{appendix:higherd}.

\subsection{Regularizing By Local Dimension $d$}

\begin{figure}[b]
\includegraphics[width=0.65\columnwidth]{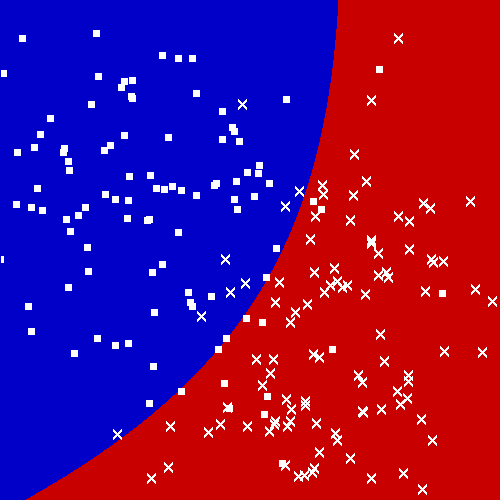}
\caption{Training points sampled from multivariate Gaussian distributions 
$P_A(x_1,x_2)$ [crosses] and $P_B(x_1,x_2)$ [squares]. The curve separating the
red $A$ region from the blue $B$ region is the theoretically optimal decision boundary.}
\label{fig:overlapping}
\end{figure}

To understand how the flexibility of the model grows with increasing $d$, consider
the case where $P_A$ and $P_B$ are overlapping distributions. Specifically, 
we take each to be a multivariate Gaussian centered respectively in the lower-right and 
upper-left of the unit square, and to have different covariance 
matrices. In Fig.~\ref{fig:overlapping} we show the theoretically optimal decision boundary
that best separates $A$ points (crosses, red region) from $B$ points (squares,
blue region), defined by the condition \mbox{$P_A(x_1,x_2)=P_B(x_1,x_2)$}.
To make a training set, we sample 100 points from each of the two distributions.

\begin{figure}[t]
\includegraphics[width=0.7\columnwidth]{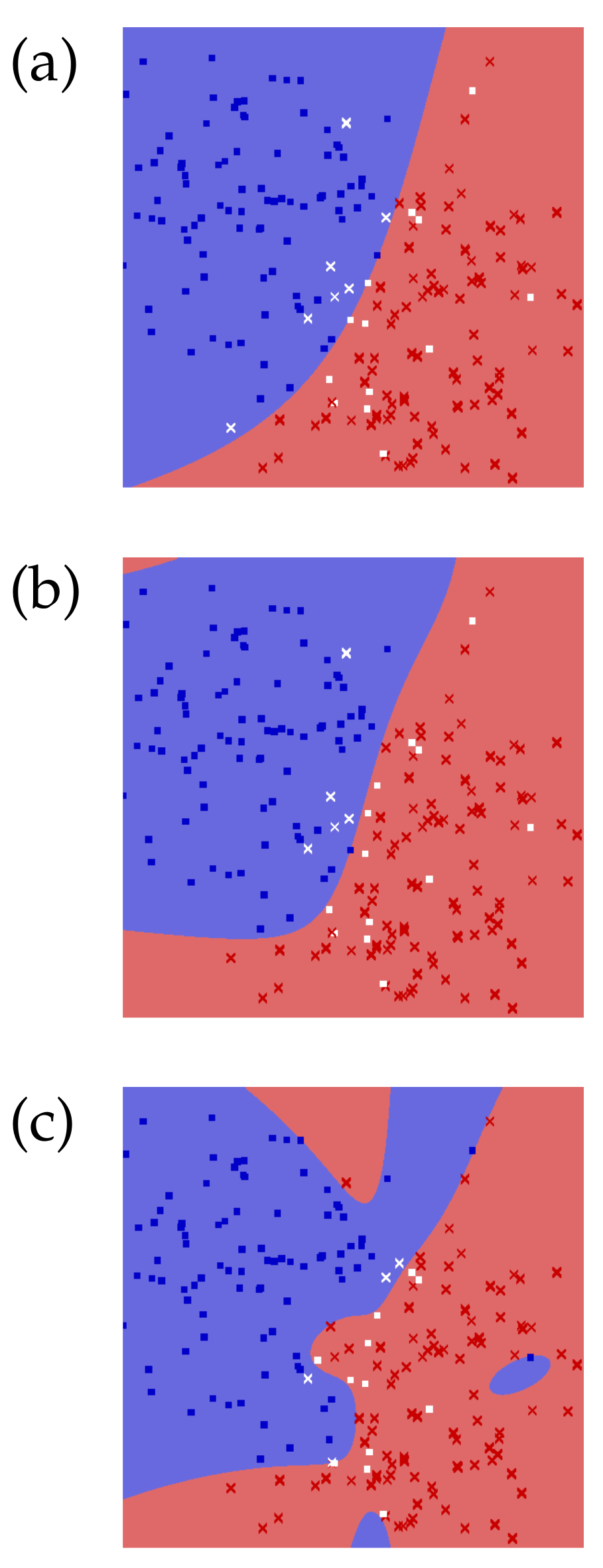}
\caption{Toy models learned from the overlapping data set Fig.~\ref{fig:overlapping}.
The results shown are for local dimension (a)~\mbox{$d=2$}, (b)~\mbox{$d=3$}, and (c)~\mbox{$d=6$}.
Background colors show how every spatial point would be classified.
Misclassified data points are colored white.}
\label{fig:dcomparison}
\end{figure}

Next, we optimize the toy model for our overlapping training set for various $d$.
The decision boundary learned by the $d=2$ model in Fig.~\ref{fig:dcomparison}(a) shows good agreement with
the optimal one in Fig.~\ref{fig:overlapping}. Because the two sets are non-separable and
this model is apparently well regularized, some of the training points are necessarily
misclassified---these points are colored white in the figure.

The \mbox{$d=3$} decision boundary shown in Fig.~\ref{fig:dcomparison} begins to show evidence of overfitting.
The boundary is more complicated than for \mbox{$d=2$} and further from the optimal boundary. Finally, 
for a much larger local dimension \mbox{$d=6$} there is extreme overfitting. The decision boundary 
is highly irregular and is more reflective of the specific sampled points than the underlying
distribution. Some of the overfitting behavior reveals the structure of the model; at the 
bottom and top of Fig.~\ref{fig:dcomparison}(c) there are lobes of one color protruding into the other.
These likely indicate that the finite local dimension still somewhat regularizes the model; otherwise
it would be able to overfit even more drastically by just surrounding each point with a small patch of its correct color.

\subsection{Non-Linear Decision Boundary}

\begin{figure}
\includegraphics[width=\columnwidth]{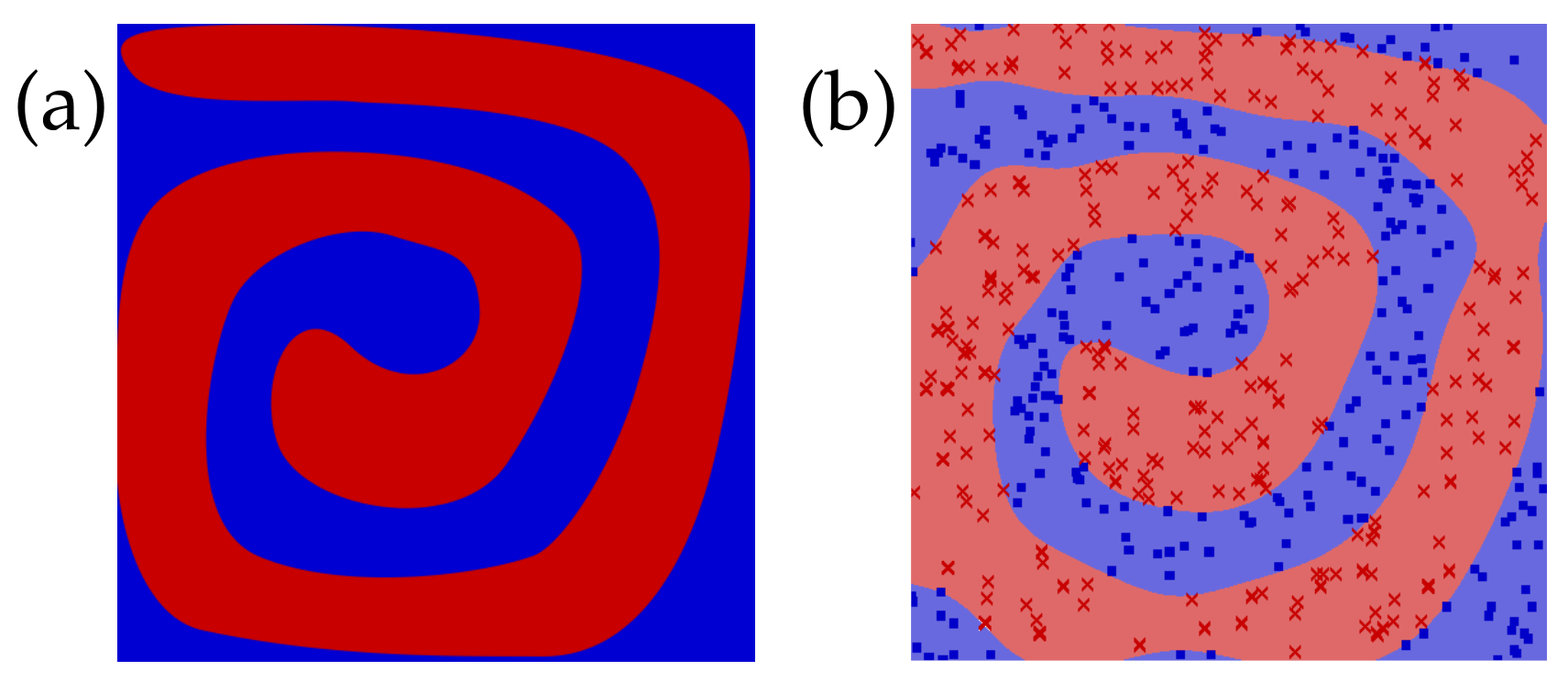}
\caption{Toy model reconstruction of interlocking spiral-shaped distribution:
(a) original distribution and (b) sampled points and distribution learned by 
model with local dimension \mbox{$d=10$}.}
\label{fig:spiral}
\end{figure}

To test the ability of our proposed class of models to learn highly non-linear decision boundaries,
consider the spiral shaped boundary in Fig.~\ref{fig:spiral}(a). Here we take $P_A$ and $P_B$ to
be non-overlapping with $P_A$ uniform on the red region and $P_B$ uniform on the blue region.

In Fig.~\ref{fig:spiral}(b) we show the result of training
a model with local dimension $d=10$ on 500 sampled points, 250 for each region 
(crosses for region $A$, squares for region $B$).
The learned model is able to classify every training point correctly, 
though with some overfitting apparent near regions with too many or too few sampled points.

\section{Interpreting Tensor Network Models}

A natural question is which set of functions of the 
form $f^\ell(\mathbf{x}) = W^\ell \cdot \Phi(\mathbf{x})$ 
can be realized when using a tensor-product feature map $\Phi(\mathbf{x})$
of the form Eq.~(\ref{eqn:tensor_prod}) and a tensor-network
decomposition of $W^\ell$. As we will argue, the possible set of functions is quite
general, but taking the tensor network structure into account provides 
additional insights, such as determining which features the model 
actually uses to perform classification.

\subsection{Representational Power}

To simplify the question of which decision functions can be realized
for a tensor-product feature map of the form Eq.~(\ref{eqn:tensor_prod}), 
let us fix $\ell$ to a single label and omit it from the notation.
We will also consider $W$ to be a completely general order-$N$ tensor with 
no tensor network constraint.
Then $f(\mathbf{x})$ is a function of the form
\begin{align}
f(\mathbf{x}) = \sum_{\{s\}} W_{s_1 s_2 \cdots s_N} \phi^{s_1}(x_1) \otimes \phi^{s_2}(x_2) \otimes \cdots \phi^{s_N}(x_N) \:.
\end{align}
If the functions $\{\phi^{s}(x)\}$, $s=1,2,\ldots,d$ form a basis for a Hilbert space of functions over 
\mbox{$x \in [0,1]$}, then the tensor product basis
\begin{align}
\phi^{s_1}(x_1) \otimes \phi^{s_2}(x_2) \otimes \cdots \phi^{s_N}(x_N)
\end{align}
forms a basis for a Hilbert space of functions 
over \mbox{$\mathbf{x} \in [0,1]^{\times N}$}. Moreover, if the basis $\{\phi^{s}(x)\}$ is complete, then the tensor product
basis is also complete and $f(\mathbf{x})$ can be any square integrable function.

Next, consider the effect of restricting the local dimension to $d=2$ as in the local feature
map of Eq.~(\ref{eqn:gray_mapping}) which was used to classify grayscale images in 
our MNIST benchmark in Section~\ref{sec:mnist}. Recall that for this choice 
of $\phi(x)$, 
\begin{align}
\phi(0) &= [1,\ 0] \\
\phi(1) &= [0,\ 1] \ .
\end{align}
Thus if $\hat{\mathbf{x}}$ is a black and white image with pixel values of only $\hat{x}_j = \{0, 1\}$,
then $f(\hat{\mathbf{x}})$ is equal to a single component $W_{s_1 s_2 \ldots s_N}$ of the weight
tensor. Because each of these components is an independent
parameter (assuming no further approximation of $W$), $f(\hat{\mathbf{x}})$ is a highly 
non-linear, in fact arbitrary, function when restricted to these black and white images.

Returning to the case of grayscale images $\mathbf{x}$ with pixels $x_j \in [0,1]$, 
$f(\mathbf{x})$ cannot be an arbitrary function over this larger space of images for finite $d$.
For example, if one considers the $d=2$ feature map Eq.~(\ref{eqn:gray_mapping}),
then when considering the dependence of $f(\mathbf{x})$ on only a single pixel $x_j$ (all other
pixels being held fixed), it has the functional form $a \cos(\pi/2 \, x_j) + b \sin(\pi/2 \, x_j)$ where $a$ and
$b$ are constants determined by the (fixed) values of the other pixels.

\subsection{Implicit Feature and Kernel Selection}

\begin{figure}[t]
\includegraphics[width=\columnwidth]{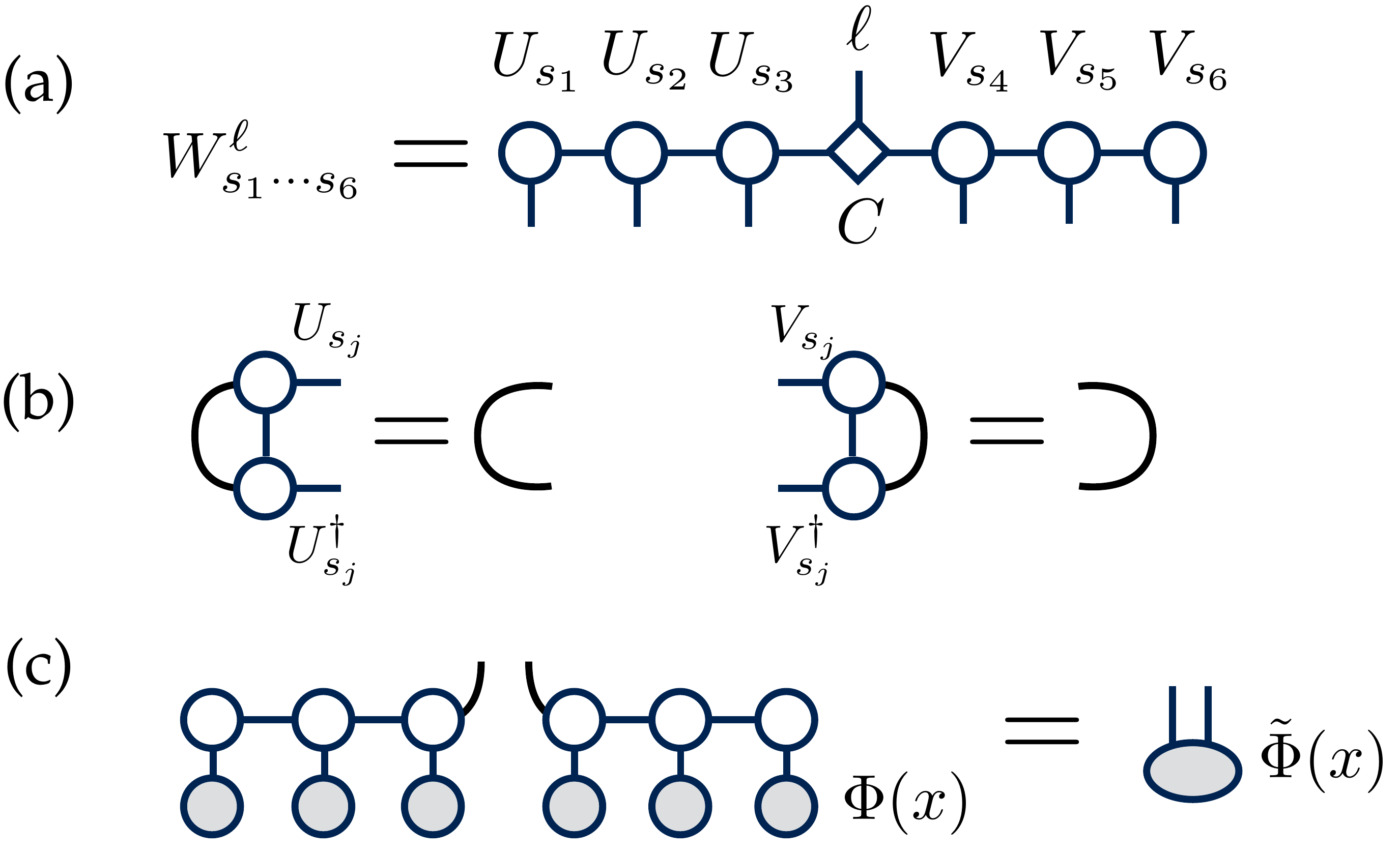}
\caption{(a) Decomposition of $W^\ell$ as an MPS with a central tensor and orthogonal site tensors. (b) Orthogonality conditions for $U$ and $V$
type site tensors. (c) Transformation defining a reduced feature map $\tilde{\Phi}(\mathbf{x})$.}
\label{fig:ortho_mps}
\end{figure}

Of course we have not been considering an arbitrary weight tensor $W^\ell$ but
instead approximating the weight tensor as an MPS tensor network. 
The MPS form implies that the decision function $f^\ell(\mathbf{x})$ has 
 interesting additional structure. One way to analyze this structure is to separate the MPS into a central tensor, or
core tensor $C^{\alpha_i \ell \alpha_{i+1}}$  on some bond $i$ and constrain all MPS site tensors to be 
\emph{left orthogonal} for sites $j \leq i$ or \emph{right orthogonal} for sites~$j \geq i$.
This means $W^\ell$ has the decomposition
\begin{align}
W^\ell_{s_1 s_2 \cdots s_N} & = \nonumber \\
 \sum_{\{\alpha\}} & \, U^{\alpha_1}_{s_1} \cdots U_{\alpha_{i-1} s_i}^{\alpha_i} C^{\ell}_{\alpha_i \alpha_{i+1}} V_{s_{i+1} \alpha_{i+2}}^{\alpha_{i+1}} \cdots V^{\alpha_{N-1}}_{s_N} \label{eqn:ortho_mps}
\end{align}
as illustrated in Fig.~\ref{fig:ortho_mps}(a). To say the $U$ and $V$ tensors are left or right orthogonal
means when viewed as matrices $U_{\alpha_{j-1} s_j}\,^{\alpha_{j}}$ and $V^{\alpha_{j-1}}\,_{s_j \alpha_{j}}$ these tensors have the property 
$U^\dagger U = I$ and $V V^\dagger = I$ where $I$ is the identity; these orthogonality conditions can be understood more clearly in terms of the diagrams
in Fig.~\ref{fig:ortho_mps}(b). Any MPS can be  brought into the form Eq.~(\ref{eqn:ortho_mps}) through an efficient sequence of tensor contractions
and SVD operations similar to the steps in Fig.~\ref{fig:sweeping2}(b).

The form in Eq.~(\ref{eqn:ortho_mps}) suggests an interpretation where the decision function $f^\ell(\mathbf{x})$ acts in three stages. 
First, an input $\mathbf{x}$ is mapped into the exponentially larger feature space defined by $\Phi(\mathbf{x})$. 
Next, the $d^N$ dimensional feature vector $\Phi$ is mapped into a much smaller
$m^2$ dimensional space by contraction with all the $U$ and $V$ site tensors of the MPS. 
This second step defines a new feature map $\tilde{\Phi}(\mathbf{x})$ with $m^2$ components
as illustrated in Fig.~\ref{fig:ortho_mps}(c). 
Finally, $f^\ell(\mathbf{x})$ is computed by contracting $\tilde{\Phi}(\mathbf{x})$ with $C^\ell$.

To justify calling $\tilde{\Phi}(\mathbf{x})$ a feature map, it follows from the left-\mbox{} and right-orthogonality
conditions of the $U$ and $V$ tensors of the MPS Eq.~(\ref{eqn:ortho_mps}) that the indices $\alpha_i$ and $\alpha_{i+1}$
of the core tensor $C$ label an orthonormal basis for a subspace of the original feature space. The vector $\tilde{\Phi}(\mathbf{x})$
is the projection of $\Phi(\mathbf{x})$ into this subspace.

The above interpretation implies that training an MPS model  
uncovers a relatively small set of important features and simulatenously learns a decision function based only on these 
reduced features. This picture is similar to popular interpretations of the hidden and output layers of
shallow neural network models \cite{Nielsen:2015}. A similar interpretation of an MPS as learning features
was first proposed in Ref.~\onlinecite{Bengua:2015a}, though with quite a
different scheme for representing data than what is used here.
It is also interesting to note that an interpretation of the $U$ and $V$ tensors as combining
and projecting features into only the $m$ most important combinations can be applied at any bond of the MPS. 
For example, the tensor $U^{\alpha_{j+1}}_{\alpha_j s_j}$ tensor at site $j$ can be viewed as defining a vector of $m$ features labeled by 
$\alpha_{j+1}$ by forming linear combinations of products of the features $\phi^{s_j}(x_j)$ and the features $\alpha_{j}$ defined by the 
previous $U$ tensor, similar to the contraction in Fig.~\ref{fig:sweeping2}(c).

\subsection{Generative Interpretation}

Because MPS were originally introduced to represent wavefunctions of quantum systems \cite{Ostlund:1995},
it is tempting to interpret a decision function $f^\ell(\mathbf{x})$ with an MPS weight vector as a wavefunction. 
This would mean interpreting $|f^\ell(\mathbf{x})|^2$ for each fixed $\ell$ as a probability distribution function
over the set of inputs $\mathbf{x}$ belonging to class~$\ell$. A major  motivation for this interpretation would 
be that many insights from physics
could be applied to machine learned models. For example, tensor networks in the same family as MPS, when
viewed as defining a probability distribution, can be used to efficiently perform perfect
sampling of the distribution they represent \cite{Ferris:2012}. 

Let us investigate the properties of $W^\ell$ and $\Phi(\mathbf{x})$ required
for a consistent interpretation of $|f^\ell(\mathbf{x})|^2$ as a probability distribution.
For $|f^\ell(\mathbf{x})|^2$  to behave like a probability distribution for a broad class of models,
we require for some integration measure $d\mu(x)$ that the distribution is normalized as
\begin{align}
\sum_\ell \int_{\mathbf{x}} |f^\ell(\mathbf{x})|^2 d\mu(x) = 1 \label{eqn:fnorm}
\end{align}
no matter what weight vector $W^\ell$ the model has, as long as the weights are 
 normalized as
\begin{align}
\sum_\ell \sum_{s_1, s_2, \ldots, s_N} \bar{W}^\ell_{s_1 s_2 \cdots s_N} W^\ell_{s_1 s_2 \cdots s_N} = 1 \ .
\end{align}
This condition is automatically satisfied for tensor-product feature maps $\Phi(\mathbf{x})$ of the form Eq.~(\ref{eqn:tensor_prod})
if the constituent local maps $\phi^s(x)$ have the property
\begin{align}
\int_x \bar{\phi}^s(x) \phi^{s^\prime}(x) \,d\mu(x) = \delta_{s s^\prime} , \label{eqn:orthophi}
\end{align}
that is, if the components of $\phi^s$ are orthonormal functions with respect to the measure $d\mu(x)$.
Furthermore, if one wants to demand, after mapping to feature space, that any input $\mathbf{x}$ itself defines a normalized distribution,
then we also require the local vectors to be normalized as
\begin{align}
\sum_s |\phi^s(x)|^2 = 1 \label{eqn:normphi}
\end{align}
for all $x \in [0,1]$. 

Unfortunately neither the local feature map Eq.~(\ref{eqn:gray_mapping}) nor its generalizations
in Appendix~\ref{appendix:higherd} meet the first criterion Eq.~(\ref{eqn:orthophi}).
A different choice that satisfies both the orthogonality condition Eq.~(\ref{eqn:orthophi}) and 
normalization condition Eq.~(\ref{eqn:normphi}) could be
\begin{align}
\phi(x) = \big[\cos(\pi x), \  \sin(\pi x) \big] \ .
\end{align}
However, this map is not suitable for inputs like grayscale pixels since it is anti-periodic
over the interval $x \in [0,1]$ and would lead to a periodic probability distribution.
An example of an orthogonal, normalized map which is not periodic on $x \in [0,1]$ is 
\begin{align}
\phi(x) = \Big[e^{i (3\pi/2) x} \cos\Big(\frac{\pi}{2} x\Big), \  e^{-i (3\pi/2) x}  \sin\Big(\frac{\pi}{2} x\Big) \Big] \ .
\label{eqn:nonperiodic}
\end{align}
This local feature map meets the criteria Eqs.~(\ref{eqn:orthophi}) and (\ref{eqn:normphi}) 
if the integration measure chosen to be \mbox{$d\mu(x) = 2 dx$}.

\begin{figure}[t]
\includegraphics[width=0.85\columnwidth]{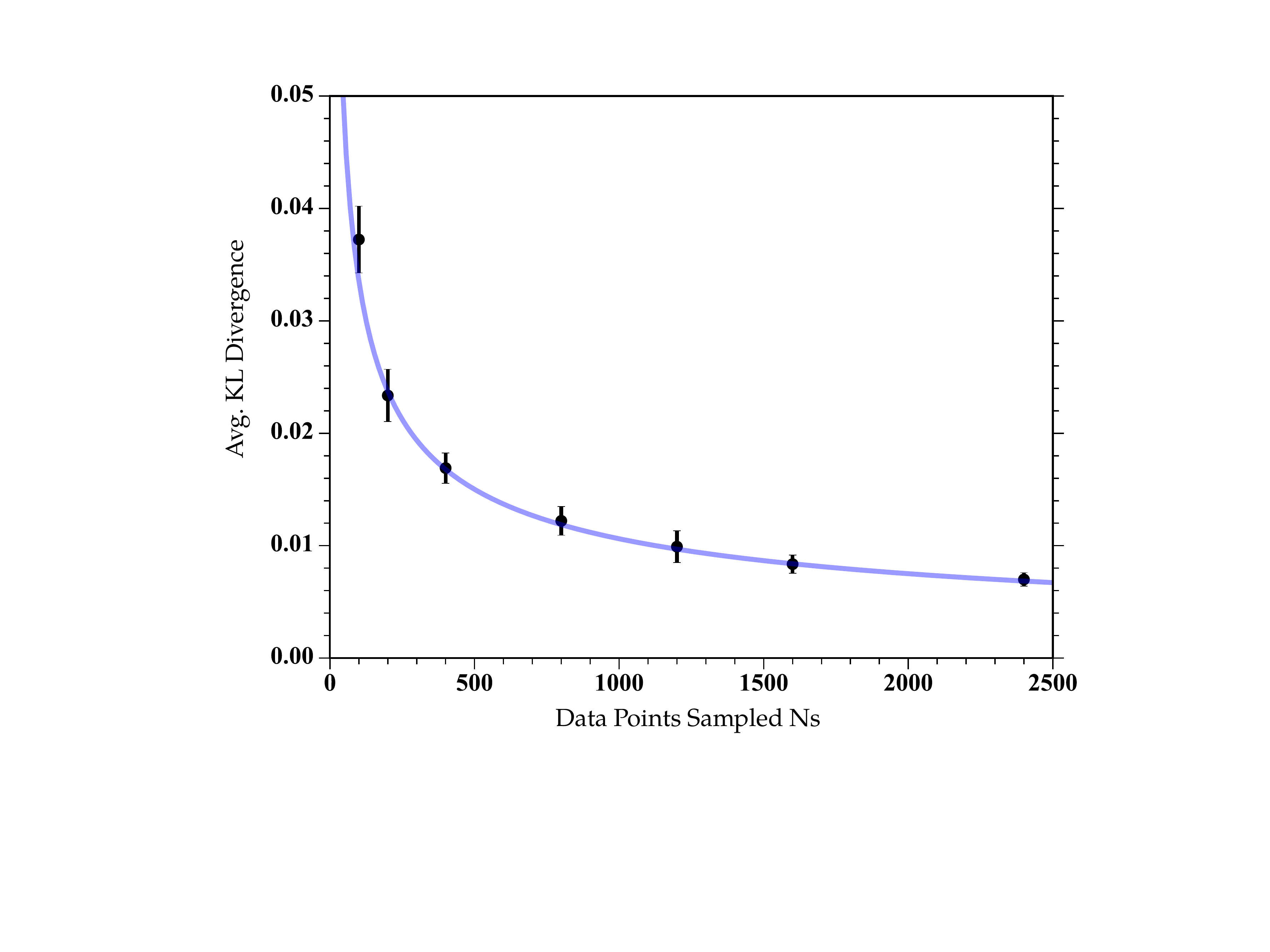}
\caption{Average KL divergence between the learned model and original model used
to generate data for a two-dimensional toy system as a function of number of sampled training points $N_s$.
The solid curve is a fit of the form $\sigma/\sqrt{N_s}$.}
\label{fig:relearn}
\end{figure}

As a basic consistency check of the above generative interpretation, we  performed an 
experiment on our toy model of Section \ref{sec:toy}, using the local feature map
Eq.~(\ref{eqn:nonperiodic}). Recall that our toy data can have two possible labels $A$ and $B$. 
To test the generative interpretation, we first generated a single, random ``hidden''
weight tensor $W$. From this weight tensor we sampled $N_s$ data points in a two step process:
\begin{enumerate}
\item Sample a label $\ell=A$ or $\ell=B$ according to the probabilities $P_A = \int_\mathbf{x} |f^A(\mathbf{x})|^2 = \sum_{s_1 s_2} |W^A_{s_1 s_2}|^2$ 
and \mbox{$P_B = 1-P_A$}.
\item Sample a data point $\mathbf{x}=(x_1, x_2)$ according to the distribution $p(\mathbf{x}|\ell) = |f^\ell(\mathbf{x})|^2/P_\ell$
for the chosen $\ell$.
\end{enumerate}
For each collection of sampled points we then trained a toy model with weight tensor $\tilde{W}$ 
using the log-likelihood cost function
\begin{align}
C = - \sum_{n=1}^{N_s} \log{|f^{L_n}(\mathbf{x}_n)|^2}
\end{align}
where recall $L_n$ is the known correct label for training point $n$.

We repeated this procedure multiple times for various sample sizes $N_s$, each time
computing the Kullback-Liebler divergence of the learned versus exact distribution
\begin{align}
D_\text{KL} = \sum_\ell \int_{\mathbf{x}} p(\ell,\mathbf{x}) \log{\!\Big(\frac{p(\ell,\mathbf{x})}{\tilde{p}(\ell,\mathbf{x})}\Big)}
\end{align}
where $p(\ell,\mathbf{x}) = |f^\ell(\mathbf{x})|^2 = |W^\ell \cdot \Phi(\mathbf{x})|^2$ and 
 $\tilde{p}(\ell,\mathbf{x})$ has similar definition in terms of $\tilde{W}$.
The resulting average $D_\text{KL}$ as a function of number of sampled training points $N_s$ is shown
in Fig.~\ref{fig:relearn} along with a fit of the form $\sigma/\sqrt{N_s}$ where $\sigma$ is a fitting
parameter. The results indicate that given enough training data, the learning process
can eventually recapture the original probabilistic model that generated the data.

\section{Discussion}
We have introduced a framework for applying quantum-inspired tensor networks to multi-class supervised learning tasks. While using an MPS ansatz 
for the model parameters worked well even for the two-dimensional data in our MNIST experiment, other tensor networks such as PEPS, 
which are explicitly designed for two-dimensional systems, may be more suitable and offer superior performance. 
Much work remains to determine the best tensor network  for a given domain.

Representing the parameters of our model by a tensor network has many useful and interesting implications. It allows one to work with a 
family of non-linear kernel learning models with a cost that is linear in the training set size for optimization, and independent of training set size for evaluation, despite using a very expressive feature map (recall in our setup, the dimension of feature space is exponential in the size of the input space).
There is much room to improve the optimization algorithm we described, adopting it to incorporate standard tricks such as 
mini-batches, momentum, or adaptive learning rates. It would be especially interesting to investigate unsupervised techniques 
for initializing the tensor network.  

Additionally, while the tensor network
parameterization of a model clearly regularizes it in the sense of reducing the number of parameters, it would be helpful to understand 
the consquences of this regularization for specific learning tasks. It could also be fruitful to include standard regularizations
of the parameters of the tensor network, such as weight decay or $L_1$ penalties. We were surprised to find good generalization without using explicit parameter regularization. For issues of interpretability, the fact that tensor networks are composed only of linear operations could be extremely useful.
For example, it is straightforward to determine directions in feature space which are orthogonal to (or projected to zero by) the weight tensor $W$.

There exist tensor network coarse-graining approaches for purely classical systems \cite{Efrati:2014,Evenbly:2014}, which could possibly be 
used instead of our approach. However, mapping the data into an extremely high-dimensional Hilbert space is likely advantageous
for producing models sensitive to high-order correlations among features. We believe there is great promise in investigating
the power of quantum-inspired tensor networks for many other machine learning tasks.\\

\textit{Note}: while preparing our final manuscript, Novikov et al. \cite{Novikov:2016} published a related framework
for parameterizing supervised learning models with MPS (tensor trains).

\subsection*{Acknowledgments} 
We would like to acknowledge helpful discussions with Juan~Carrasquilla, Josh~Combes, Glen~Evenbly, Bohdan~Kulchytskyy, 
Li~Li, Roger~Melko, Pankaj~Mehta, U.~N.~Niranjan, Giacomo~Torlai, and Steven~R.~White.
This research was supported in part by the \href{http://perimeterinstitute.ca}{Perimeter Institute for Theoretical Physics}. Research at Perimeter Institute is supported by the Government of Canada through Industry Canada and by the Province of Ontario through the Ministry of Economic Development \& Innovation.
This research was also supported in part by the Simons Foundation Many-Electron Collaboration.

\appendix

\section{Graphical Notation for Tensor Networks \label{appendix:graphical}}

Though matrix product states (MPS) have a relatively simple structure,
more powerful tensor networks, such as PEPS and MERA, have such complex structure
that traditional tensor notation becomes unwieldy. For these networks, and even
for MPS, it is helpful to use a graphical notation. For some more complete reviews of this 
notation and its uses in various tensor networks see Ref.~\onlinecite{Bridgeman:2016,Cichocki:2014b}.

\begin{figure}
\includegraphics[width=0.7\columnwidth]{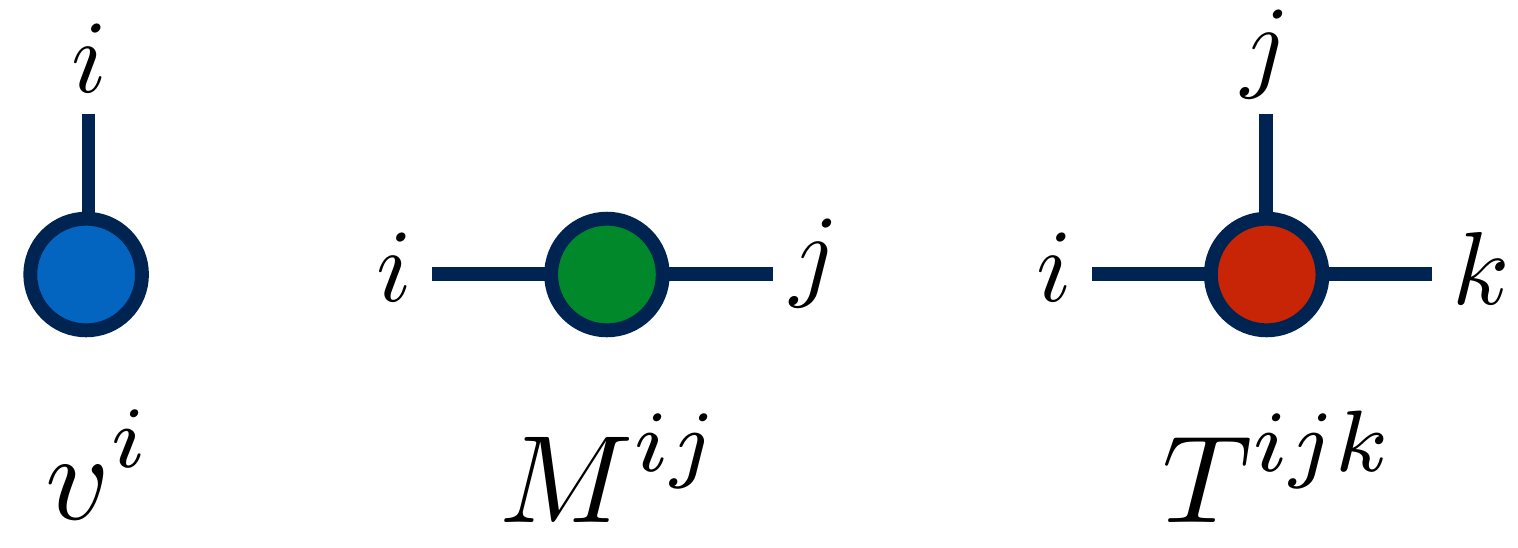}
\caption{Graphical tensor notation for (from left to right) a vector, matrix, and order 3 tensor.}
\label{fig:graphical}
\end{figure}

The basic graphical notation for a tensor is to represent it as a closed shape. 
Typically this shape is a circle, though other shapes can be used to distinguish 
types of tensors (there is no standard convention for the choice of shapes).
Each index of the tensor is represented by a line emanating from it; 
an order-N tensor has N such lines. Figure~\ref{fig:graphical} shows examples of diagrams 
for tensors of order one, two, and three.

\begin{figure}
\includegraphics[width=0.6\columnwidth]{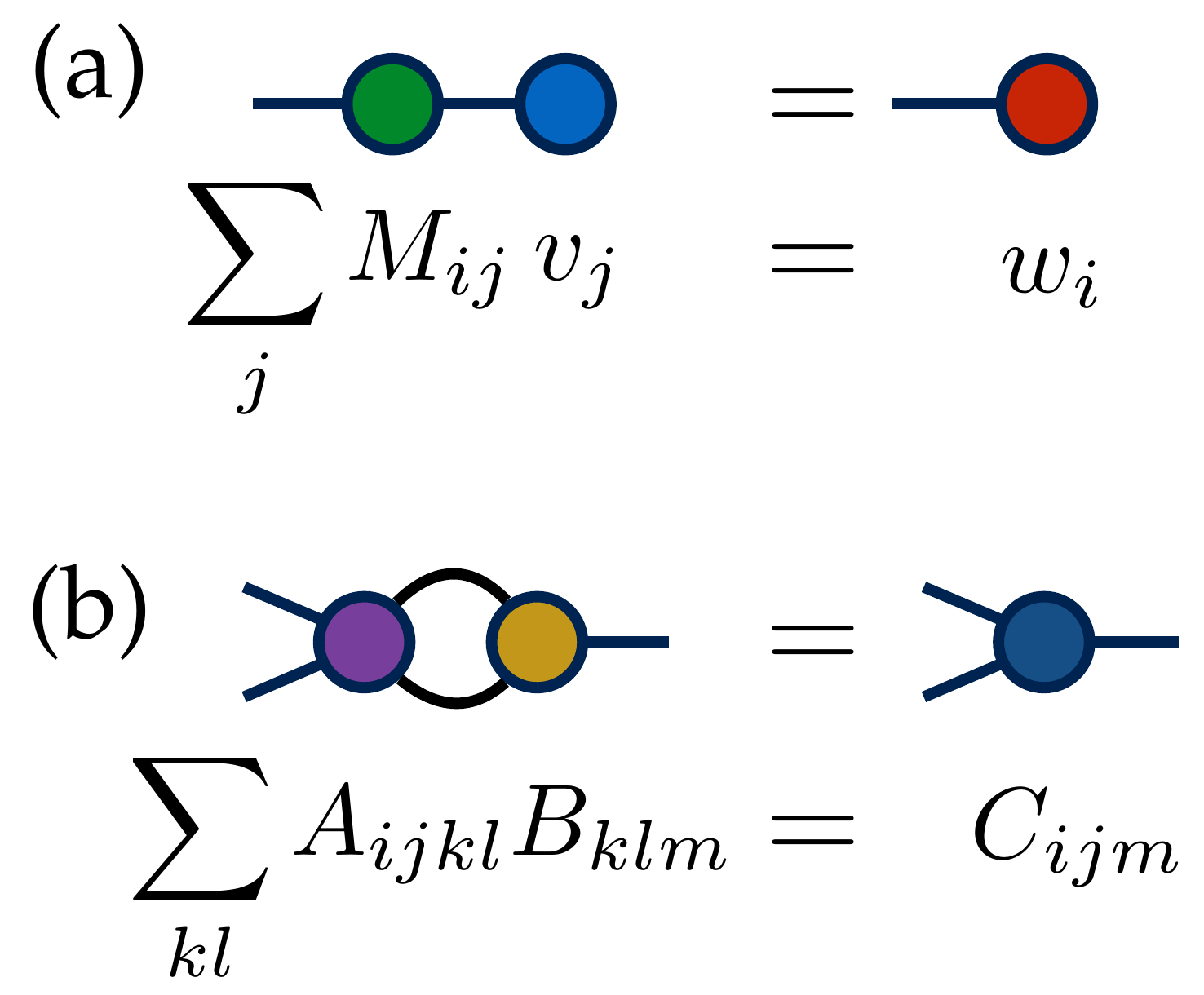}
\caption{Tensor diagrams for (a) a matrix-vector multiplication and (b) a more general tensor contraction.}
\label{fig:contract}
\end{figure}

To indicate that a certain pair of tensor indices are contracted, 
they are joined together by a line. For example, Fig.~\ref{fig:contract}(a) shows the contraction 
of an an order-1 tensor with the an order-2 tensor; this is the usual
matrix-vector multiplication.
Figure~\ref{fig:contract}(b) shows a more general
contraction of an order-4 tensor with an order-3 tensor.

Graphical tensor notation offers many advantages over traditional notation. 
In graphical form, indices do not usually require names or labels since they can be distinguished
by their location in the diagram. Operations such as the outer product,
tensor trace, and tensor contraction can be expressed without additional
notation; for example, the outer product is just the placement of one tensor next to another. 
For a network of contracted tensors, the order of the final resulting tensor
can be read off by simply counting the number of unpaired lines left over. 
For example, a complicated set of tensor contractions can be recognized as giving a scalar result
if no index lines remain unpaired.

Finally, we note that a related notation for sparse or structured matrices in a direct-sum formalism can 
be used, and appears extensively in Ref.~\onlinecite{Fishman:2015}.

\section{Higher-Dimensional Local Feature Map \label{appendix:higherd}}

As discussed in Section~\ref{sec:encoding}, our strategy for using tensor networks to classify input data
begins by mapping each component $x_j$ of the input data vector $\mathbf{x}$ to a $d$-component
vector $\phi^{s_j}(x_j)$, $s_j = 1,2,\ldots,d$. We always choose $\phi^{s_j}(x_j)$ to be a unit vector
in order to apply physics techniques which typically assume normalized wavefunctions. 

For the case of $d=2$ we have used the mapping 
\begin{equation}
\phi^{s_j}(x_j) = \Big[\cos\Big(\frac{\pi}{2} x_j\Big),\,\sin\Big(\frac{\pi}{2} x_j\Big)\Big] \:. \label{eqn:d2}
\end{equation}
A straightforward way to generalize this mapping to larger $d$ is as follows.
Define $\theta_j=\frac{\pi}{2} x_j$.
Because $(\cos^2(\theta_j)+\sin^2(\theta_j))=1$, then also
\begin{equation}
(\cos^2(\theta_j)+\sin^2(\theta_j))^{d-1}=1 \ .
\end{equation}
Expand the above identity using the binomial coeffiecients $\binom{n}{k} = n!/(k!(n-k)!)$.
\begin{align}
\MoveEqLeft (\cos^2(\theta_j)+\sin^2(\theta_j))^{d-1} = 1 \nonumber \\
& = \sum_{p=0}^{d-1} \binom{d-1}{p} (\cos\theta_j)^{2(d-1-p)} (\sin\theta_j)^{2p} \:.
\end{align}
This motivates defining $\phi^{s_j}(x_j)$ to be
\begin{equation}
\phi^{s_j}(x_j) = \sqrt{\binom{d-1}{s_j-1}}\ (\cos(\frac{\pi}{2} x_j))^{d-s_j} (\sin(\frac{\pi}{2} x_j))^{s_j-1}
\end{equation}
where recall that $s_j$ runs from $1$ to $d$. The above definition reduces to the $d=2$ case Eq.~(\ref{eqn:d2}) and
guarantees that $\sum_{s_j} |\phi^{s_j}|^2 = 1$ for larger $d$. (These functions are
actually a special case of what are known as \emph{spin coherent states}.)

Using the above mapping $\phi^{s_j}(x_j)$ for larger $d$ allows the product
$W^\ell \cdot \Phi(\mathbf{x})$ to realize a richer class of functions. One reason is
that a larger local dimension
allows the weight tensor to have many more components. 
Also, for larger $d$ the 
components of $\phi^{s_j}(x_j)$ contain ever higher frequency 
terms of the form $\cos\big(\frac{\pi}{2} x_j\big)$, $\cos\big(\frac{3 \pi}{2} x_j\big)$, \ldots, 
$\cos\big(\frac{(d-1) \pi}{2} x_j\big)$ and similar terms replacing $\cos$ with $\sin$.
The net result is that the decision functions realizable for larger $d$ are composed
from a more complete basis of functions and can respond to smaller variations in $\mathbf{x}$.

\bibliography{mpsml}

\end{document}